# Predicting Diabetes Using Machine Learning: A Comparative Study of Classifiers

Mahade Hasan[1]*, Farhana Yasmin[1]

[1]*Nanjing University of information Science and Technology, Nanjing China.*
*Corresponding Author: mhasan@nuist.edu.cn*

**ABSTRACT-** Diabetes remains a significant health challenge globally, contributing to severe complications like kidney disease, vision loss, and heart issues. The application of machine learning (ML) in healthcare enables efficient and accurate disease prediction, offering avenues for early intervention and patient support. Our study introduces an innovative diabetes prediction framework, leveraging both traditional ML techniques such as Logistic Regression, SVM, Naïve Bayes, and Random Forest and advanced ensemble methods like AdaBoost, Gradient Boosting, Extra Trees, and XGBoost. Central to our approach is the development of a novel model, DNet, a hybrid architecture combining Convolutional Neural Network (CNN) and Long Short-Term Memory (LSTM) layers for effective feature extraction and sequential learning. The DNet model comprises an initial convolutional block for capturing essential features, followed by a residual block with skip connections to facilitate efficient information flow. Batch Normalization and Dropout are employed for robust regularization, and an LSTM layer captures temporal dependencies within the data. Using a Kaggle-sourced real-world diabetes dataset, our model evaluation spans cross-validation accuracy, precision, recall, F1 score, and ROC-AUC. Among the models, DNet demonstrates the highest efficacy with an accuracy of 99.79% and an AUC-ROC of 99.98%, establishing its potential for superior diabetes prediction. This robust hybrid architecture showcases the value of combining CNN and LSTM layers, emphasizing its applicability in medical diagnostics and disease prediction tasks.

## 1. INTRODUCTION

Hyperglycemia is a hallmark of the clinical illness known as diabetes mellitus, which has multiple etiologies (mellitus being Latin for sweet). About 90% of those with diabetes have type 2, with the rest having type 1 [1]. Over 422 million people worldwide suffer from diabetes, the majority of whom reside in low- and middle-income nations[2]. By keeping a healthy weight and managing blood sugar levels, diabetes can be avoided. Every year, diabetes directly causes the deaths of around 1.5 million people around the world. Throughout the past few decades, there has been an increase in the prevalence of diabetes overall as well as a steady increase in the number of cases of the condition overall [3].The long-term illness of diabetes has an impact on how the body uses food as fuel. The majority of the food you eat is transformed by the body into sugar (glucose), which is then absorbed into your bloodstream. The pancreas secretes insulin in response to an increase in blood sugar levels. In the presence of insulin, blood sugar can enter cells and be utilized as a source of energy. Insufficient or improper insulin usage by the body can result in diabetes. Too much blood sugar is left in the bloodstream when there is not enough insulin or when cells cease reacting to insulin. Severe health issues like heart disease, kidney failure, and vision loss may follow it. Global health systems serve more people [4]. Many deadly diseases impact people worldwide. Diabetes causes heart attacks, kidney failure, blindness, and other major health issues. Every hospital tracks illnesses. IT has transformed health care [5]. Each machine-learning algorithm enhances disease prediction and healthcare automation. Hadoop, built on computer clusters, efficiently analyzes and stores massive datasets in the cloud [6], [7]. This work [8] suggests forecasting diabetes with Hadoop-based machine learning. The results imply machine-learning algorithms can accurately forecast diabetes. The National Institute of Diabetes and Digestive Diseases' Pima Indians Diabetes Database examines the algorithm's efficacy. A cutting-edge computing healthcare system is one of the most investigated topics in healthcare research[9]. Computing and healthcare researchers collaborate to advance such systems [10]. The World Health Organization (WHO) found an increase in diabetic patients and deaths. Diabetes can have long-term effects. Medical research is expanding. Technological advances and continual monitoring are needed to collect, preserve, study, and forecast such individuals' health. India's rising diabetic population is worrying. A system that records, analyzes, and searches for diabetic risks using technology is essential. Researchers are developing early detection and analysis methods. This paper [11] reviews diabetic studies and prospective frameworks.

Recently, healthcare research is popular and data-driven. Healthcare data volumes require big data analytics. Millions worldwide use various treatments. If patient care patterns for a disease are examined first, conclusions will be more informed. Healthcare advancements continue to enhance overall well-being. Clinicians can use machine learning to diagnose diseases early. This WEKA-based study [12] creates a diabetes classifier. Naive Bayes, Support Vector Machine, Random Forest, and Simple CART will forecast the study. The project will recommend the best diabetes prediction algorithm. Each method's dataset was assessed. The SVM predicted diseases best. Modern diets and erratic eating habits are increasing the prevalence of diabetes. Obesity and high blood glucose levels are major diabetes risk factors. This study [13] investigates the main causes of diabetes. Variable and feature selection has become a prominent research topic in application domains with easily accessible datasets with tens or hundreds of pieces. Machine learning shows this trend. They'll also focus on key factors to consider when assessing diabetes risk.

Medical diagnostic software development is difficult due of disease prediction. Medical diagnosis is one application of machine learning. A machine learning algorithm-based classifier system may help medical personnel identify and predict diseases and solve health issues. Machine learning classification algorithms can improve a disease diagnosis system's efficiency, effectiveness, dependability, and precision. This article [14] discusses machine learning-based diabetes diagnosis. The PIMA Indian Diabetic dataset also used artificial neural networks, decision trees, random forests, naive Bayes, support vector machines, logistic regression, and k-nearest neighbors. These analyses and their pros and cons were then examined. Predictive analytics on massive data sets generally uses machine learning methods. Predictive analytics in medicine is difficult to execute, but it can help doctors make quick choices about patient health and therapy based on massive volumes of data. This study examines healthcare predictive analytics using six machine learning algorithms. The experiment uses six machine learning algorithms on patient medical records. Several methods are compared for efficiency and accuracy. The study's machine learning approaches yielded the best diabetes prediction system. This study [15] uses machine learning to help doctors diagnose diabetes early.

Moreover, healthcare generates sensitive data. Predicting diabetes is medical goal. Machine learning methods can examine data and synthesize decision-making expertise. Data mining can provide valuable insights from enormous amounts of accessible data. Analyzing new patterns gives customers vital information. Diabetes raises heart, kidney, nerve, and eye risks. Data mining will classify and identify Diabetes dataset trends. The UCI Pima Indian diabetes database was used. A sophisticated model used to predict and diagnose diabetes. Bootstrapping resampling improves accuracy in this investigation [16]. Next, Naive Bayes, Decision Trees, and KNN algorithms analyzed each strategy. Furthermore age, obesity, lack of exercise, family history, improper food, high blood pressure, etc. can cause diabetes. Diabetes increases the risk of cardiovascular disease, renal disease, stroke, vision issues, nerve damage, and more. Today, hospitals utilize many tests to detect and treat diabetes. Healthcare and medicine use big data analytics. Healthcare organizations have enormous databases. Big data analytics can scan enormous databases, find new information and trends, and make predictions. Current categorization and prediction fail. This study [17] suggests using blood glucose, body mass index, age, and insulin levels to predict diabetes. This model adds diabetes risk factors. New datasets enhance categorization. Diabetes prediction improved with pipeline models.

As diabetes mellitus, a rising chronic condition, is caused by the body's inability to metabolize glucose. Another study [18] created a prediction model with excellent sensitivity and selectivity to better identify

Canadian patients at risk of diabetes mellitus based on demographic data and laboratory findings from healthcare visits. Most Americans die from cardiovascular disease and diabetes. Recognizing and preparing for these diseases' patient presentations is the first step in halting their progression. Using survey data, [19] they uncover data elements that contribute to prevalent diseases and evaluate the capacity of machine learning models to detect at-risk patients and test results.

On the other hand, in recent years, machine learning has become an increasingly popular method for the analysis of datasets pertaining to medical subjects. This study aims to find a solution by determining the most effective model for the early prediction of diabetes through the application of techniques derived from machine learning. The following is a list of the most important ramifications that this study has:

- The data conversion for the further simulation.
- Conducted association rule mining in order to determine the most common pattern of diabetes symptoms.
- Applied Data Preprocessing on the dataset.
- Applied six different models on the dataset to find out important features.
- Applied eleven different models on the dataset.
- Comparing the performance of those models to identify the best model.
- Analyzing how the best model performed in comparison to earlier research.

The following sections are included in this study: Related Work, Experimentation Environment, Methodology, Data Collection, Data Conversion, Rule Mining, Data Preprocessing, Important Feature Selection, Train Test Split, Applied Models, Result Analysis, Discussion, Comparison with Existing Works, and Conclusion. The section that follows offers some additional explanation in light of this.

## 2. LITERATURE REVIEW

This section covered the research that only makes use of the diabetes dataset, and it included a review of the relevant literature to our study. During the reviewing those pieces of literature, took into consideration the research methodologies as well as the results of the investigations.

Tripathi et al,. [20] Diabetes impacts glucose. Insulin resistance causes difficulties. Undiagnosed, it destroys kidneys, nerves, and eyes. Technology enhances individualized medicine. Healthcare uses machine learning, a fast-growing predictive analysis subfield. These tools detect illnesses. Machine learning categorization and diabetes-related factors predict diabetes early in this study. It improves patient diagnosis and produces clinically useful results. Four ML algorithms predict early diabetes. LDA, KNN, SVM, Random Forest (RF). Pima Indian Diabetes Database from UC Irvine's machine learning repository is used in experiments (PIDD). Categorization systems are evaluated using the metrics sensitivity (recall), precision, specificity, F-score, and accuracy. Suitable categories. The accuracy of RF categorisation is 87.66%. The accuracy of RF categorisation is 87.66%.

Alaa Khaleel et al,. [21] Diabetics have high blood sugar levels. Early detection lowers diabetes severity and risk. Machine learning, especially in disease, is becoming more popular in medicine due to its ubiquity. This study models diabetes diagnosis. [Ref] they evaluate powerful machine learning (ML) algorithms' prediction precision using precision, recall, and F1-measure. The PIDD dataset predicted diabetic symptoms. LR, NB, and KNN exhibited 94%, 79%, and 69% accuracy, respectively. LR predicts diabetes better.

Zou et al,. [22] Diabetes produces hyperglycemia. It is risky. Morbidity will cause 642 million diabetes cases by 2040. Medical and public health employ machine learning. Their decision trees, random forests, and neural networks predicted diabetes. Hospital patients in Luzhou, China, are examined. This experiment cross-validated five models. They used top methodologies to evaluate their viability. High-performing approaches did this. 68994 healthy and diabetes patients were trained. Unbalanced data pulled 5 times more. Answer: five-test average. MRMR and PCA lowered this study's dimension (mRMR). Random forest prediction was best overall (ACC = 0.8084).

Tigga et al,. [23] India has approximately 30 million diabetics and many more at risk. Thus, early diagnosis and treatment are needed to prevent diabetes and its complications. This study estimates diabetes risk from lifestyle and family history. Machine learning algorithms, which are accurate, predicted type 2 diabetes risk. Medical workers need accuracy. Individuals can estimate their diabetes risk after the model is trained. For the trial, 952 participants completed an online and offline questionnaire. The 18-question assessment covers health, lifestyle, and family history. The Pima Indian Diabetes database was evaluated using the same techniques. Random Forest Classifier performs best on both datasets.

Sisodia et al,. [24] Diabetes elevates glucose (sugar). Undiagnosed diabetes can create several issues. The patient always sees a doctor at a diagnostic facility because identification takes so long. Machine learning solves this major problem. This study seeks a model that reliably predicts diabetes risk. This experiment detects early diabetes using decision trees, SVMs, and naive bayes. UC Irvine's machine learning repository studies the Pima Indians Diabetes Database (PIDD). Recall, F-measure, precision, and accuracy evaluate each approach. Incident categorization measures accuracy. Naive Bayes' 76.30% accuracy beats others. ROC curves verify this.

Ramesh et al,. [25] Millions have diabetes. It worsens organ failure and life quality. Diabetics need early detection and monitoring. Remote patient monitoring facilitates treatment. This study proposes an end-to-end remote monitoring infrastructure for automated diabetes risk prediction and management. Smartphones, smart wearables, and health devices fuel the platform. A Pima Indian Diabetes Database Support Vector Machine predicted diabetes risk after scaling, imputation, selection, and augmentation. Tenfold stratified cross validation produced 83.20% accuracy, 87.20% sensitivity, and 79% specificity. Smartphones and smartwatches measure vitals, slow diabetes, and connect with doctors. The unobtrusive, economical, and vendor-interoperable platform aids doctors in their decisions by using the most recent diabetes risk projections and lifestyle data.

Perveen et al,. [26] Interventional programs can save time and dollars by targeting high-risk diabetics. A Hidden Markov Model (HMM), a machine learning tool, was tested to evaluate the Framingham Diabetes Risk Scoring Model (FDRSM), a respected prognostic model. HMM: 8-year diabetes risk prediction? No HMM study has validated FDRSM performance. From 172,168 primary care patients' EMRs, HMM calculated 8-year diabetes risk. Our 911-person sample with all risk indicators and follow-up data had an AROC of 86.9%, compared to 78.6% and 85% in a recent FDRSM validation analysis in the same Canadian population and the Framingham research, respectively. 911 research participants with all risk indicators and follow-up data had 86.9% AROC. The recommended HMM discriminates better than the FDRSM validation research in Canada and Framingham. HMM can identify eight-year diabetes risk.

Maniruzzaman et al,. [27] Diabetes causes high blood sugar. It can cause heart attack, kidney failure, stroke, and other serious illnesses. 422 million people had diabetes in 2014. 642 million will live on Earth by 2040. This project creates an ML-based diabetes diagnosis system.

Kavakiotis et al,. [28] Biotechnology and health science have increased high-throughput genetic data and clinical information from vast electronic health records (EHRs). To assess all data, biosciences must use machine learning and data mining. DM affects global health. Diabetes diagnosis, etiopathophysiology, management, and other areas have been studied. Machine learning, data mining, and diabetes research methods will be examined for prediction, diagnosis, complications, genetic background and environment, health care, and management. Popularity is first. ML algorithms were utilized. Methods were 85% monitored, but association rules were not. Support Vector Machines (SVMs) reign supreme in their application and performance. Clinical data emerges as the cornerstone of informed decision-making in healthcare. The names of the selected publications show that extracting vital knowledge generates new ideas that improve DM comprehension and research.

Hasan et al,. [29] Diabetes leads to an elevation in glucose levels within the body. Early detection reduces diabetes risk. Detecting diabetes early decreases the risk of its onset. Outliers and unlabeled data complicate diabetes prediction. This literature's robust diabetes prediction framework used outlier rejection, filling missing values, data standardization, feature selection, K-fold cross-validation, several Machine Learning (ML) classifiers (k-nearest Neighbor, Decision Trees, Random Forest, AdaBoost, Naive Bayes, and XGBoost), and Multilayer Perceptron (MLP). Guidelines: ML Area ROC Curve weights (AUC). This study suggests weighting diabetes prediction ML models. Grid search maximizes hyperparameter adjustment AUC. This study used the Pima Indian Diabetes Dataset and identical experimental parameters. Our recommended ensembling classifier is the most successful classifier from exhaustive testing, with a diagnostic odds ratio of 66.234, an AUC of 0.950, a sensitivity of 0.789, a specificity of 0.934, a false omission rate of 0.092, and a diagnostic odds ratio of 0.934. 2.0% lower AUC. Poor diabetes prediction. Same dataset may improve diabetes prediction systems. Diabetic prognosis.

Yahyaoui et al,. [30] DSS helps doctors and nurses make clinical decisions. This is needed due to escalating deadly diseases. Diabetes is a global contributor to mortality. Raising blood sugar may influence other organs. Diabetes. The International Diabetes Federation (IDA) estimates 592 million people will have diabetes by 2035. This research proposes a machine learning-based diabetes prediction DSS. SVM and Random Forest classifiers are popular (RF). Fully convolutional neural networks (CNNs) predicted and detected diabetics (DL). The public Pima Indians Diabetes database had 768 samples with 8 features to evaluate the suggested approach. 500 samples were non-diabetic, 268 were. SVM 83.67%, RF 76.81%, and DL 65.38%. RF outperforms deep learning and SVM in diabetes prediction.

Sonar et al,. [31] diabetics causes blindness, urinary system problems, coronary heart disease, and more. After the consultation, the patient must drive to a diagnostic center for their reports, which takes time and money. Machine learning can now solve it. Polygenic disease is diagnosed using cutting-edge information processing. Anticipating illness allows for critical care. Data from a lot of unviewed diabetes-related information is removed. This study will improve diabetic risk prediction. SVM algorithms, naïve bayes networks, decision trees, and AI networks characterize models (ANN). Decision Tree, Naive Bayes, and Support Vector Machine models estimate 85%, 77%, and 77.3% precision. Results are accurate.

Sivaranjani et al,. [32] Diabetes is one of the most common and deadly diseases worldwide, including in India. Lifestyle, genetics, stress, and age can cause diabetes at any age. Untreated diabetes, regardless of cause, can have catastrophic consequences. Several methods can anticipate diabetes and its complications. Researchers employed SVM and Random Forest machine learning techniques in the suggested work (RF). These algorithms estimate diabetes risk. After data preparation, step forward and backward feature selection identifies predictive qualities. Selecting features, PCA dimensionality reduction is studied. Random Forest (RF) has an 83% prediction accuracy, compared to SVM's 81.4%.

Saha et al,. [33] Diabetes, a prevalent condition, can strike at any age. These diseases activate when blood sugar rises. Predicting diabetes is crucial right now. The Indian Pima Dataset has undergone many techniques. This dataset includes Pima Indian women's 1965 research. Most academics are trying to apply difficult methods to datasets, however a lot of in-depth research lacks easy strategies. This study included RF, SVM. They used these methods in several ways. They added several methods to the main dataset. They then identified diabetics using preprocessing methods. They compared and got the best outcomes using those methods. Neural Network was the most accurate method (80.4%).

Posonia et al,. [34] Diabetes mellitus, which can cause severe birth abnormalities, affects most Indian pregnant women. Several cutting-edge blood test technologies can detect diabetes. Diabetes results from elevated blood glucose. Untreated diabetes can cause renal damage and heart attacks. Thus, discovering and studying gestational diabetes requires learning models and rigorous research. This study suggested diabetes prediction using machine learning. Example: Decision Tree J48 calculation. "Decision Tree" is a top classification model. 768 patients with major 8 qualities and a goal column indicating positive or negative were analyzed. Our Weka experiment showed that the Decision Tree J48 calculation is more effective and faster.

Pavani et al,. [35] Today's healthcare uses AI and ML. The WHO says diabetes affects the most individuals worldwide. High glucose levels induce it. Diagnosing diabetes may involve other factors. This research aims to develop a diabetes-prediction system. This study employed ML methods to predict early diabetes. Machine learning technologies include support vector machines, logistic regression, decision trees, random forests, gradient boost, K-nearest neighbor techniques, and Naive Bayes. These algorithms are evaluated using precision, accuracy, recall, and F-measure. This study compares approaches to improve precision. The Random Forest algorithm and Naive Base Method has 80% accuracy.

Let's quickly go over the methodology that was applied in this research now that this point has been established. The section on the methodology, which will be presented after this part, will contain additional information on this topic.

## 3. METHODOLOGY

The ten distinct primary sections had to be completed in order for this study to be finished. The presentation and discussion of the specifics of the dataset's description takes up one section of the "Data Collection" section. In addition, the history of the data set has been thoroughly dissected and examined. The string data is converted into numerical data in the "Data Conversion" section. The article's section on "Rule Mining" briefly mentioned how datasets frequently correlate with one another. The report's "Data Preprocessing" section has been updated with the necessary data preprocessing techniques. Additionally, six different models have been used to determine which important features are most advantageous in the section titled "Important Feature Selection". The dataset has been divided into a train set and a test set in the "Train Test Split" section of the document so that the experiment can be run on both of them. A total of eleven different models have been used to analyze the dataset and predict the likelihood of early diabetes, and they are listed in the "Applied Model" section of the paper. The discussion of the model's performance that was chosen as having the best overall performance after evaluating all of the models has since moved on to the "Result Analysis" Section. In conclusion, the results of the model that performed the best have been analyzed and compared to those of work that has already been published in order to make an accurate prediction of early diabetes using machine learning. Figure 3-1 depicts the step-by-step procedure that was used in this investigation.

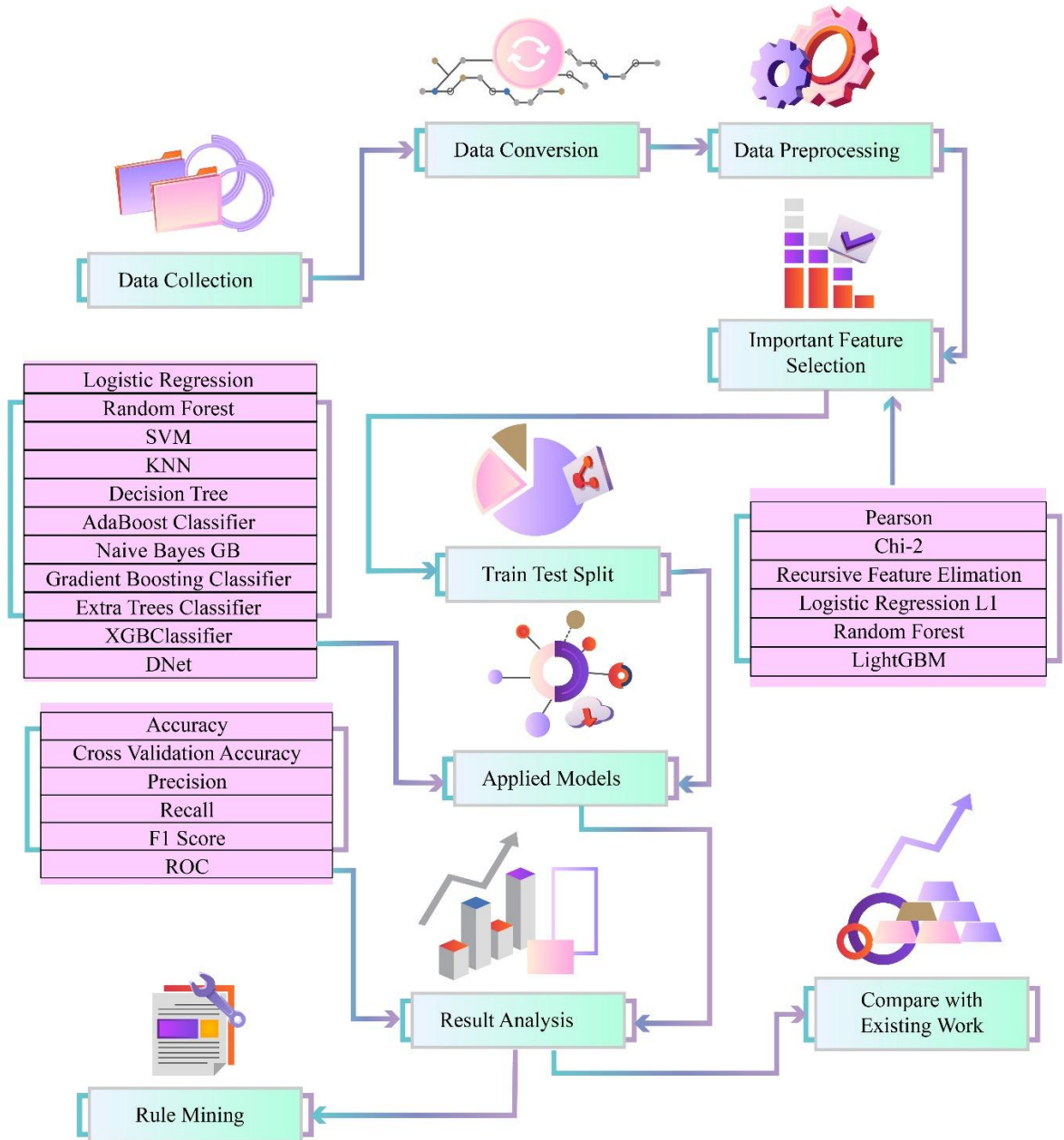

*Figure 3-1. **Illustration of working flowchart***

The dataset Section has now started the working process of this study in order to discuss the dataset's attributes in the following way.

## 4. Description of Dataset used
This dataset includes information on the signs and symptoms of diabetic patients who have recently been diagnosed or who are at risk for developing diabetes. The patients at the Sylhet Diabetes Hospital in Sylhet, Bangladesh, filled out direct questionnaires to provide this information, and a medical professional gave the project the green light before it was carried out. Diabetes is linked to 520 different patients and 16 different characteristics. There is one continuous attribute in addition to fifteen different categories of attributes. The dataset includes a total of 15 features, one of which is the target variable defined as class. Table 4-1 shows the summarized description of the dataset.

*Table 4-1 The Summarized Description of the Dataset*

| Data Set Characteristics: | Multivariate |
|---|---|
| Attribute Characteristics: | N/A |
| Associated Tasks: | Prediction |
| Number of Instances: | 520 |
| Number of Attributes: | 16 |
| Missing Values | Yes |

## 4.2 CORRELATION WITH DIABETES

The importance of the feature link with diabetes cannot be overstated when it comes to early diabetes predictions. In our data, the correlations for the following variables are as follows: age has a correlation of 0.10, gender has a correlation of -0.44, polyuria has a correlation of 0.66, polydipsia has a correlation of 0.64, sudden weight loss has a correlation of 0.43, weakness has a correlation of 0.24, polyphagia has a correlation of 0.34, genital thrush has a correlation of 0.11, visual blurring has a correlation of 0.2. Figure 4-18 is an instance of the feature correlation that has been demonstrated to be associated with diabetes.

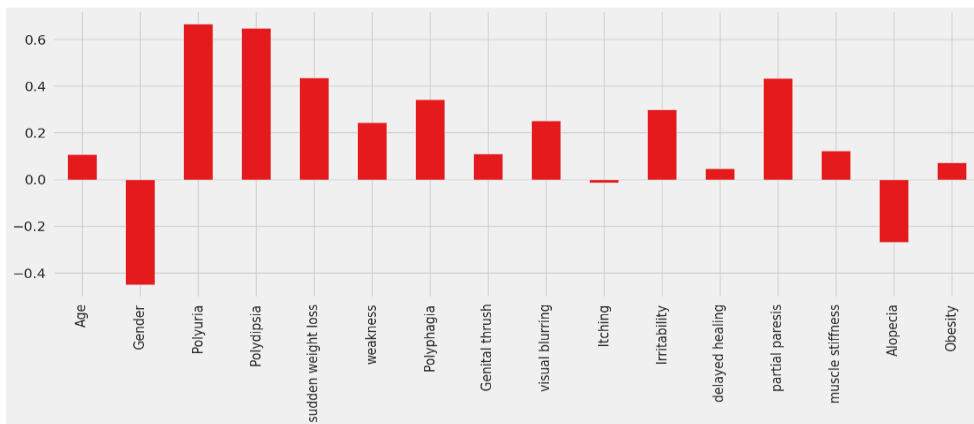

*Figure 4-18 Features Correlation with Diabetes*

The mining of association rules will now be carried out on the dataset in order to complete Mining Association Rules. Before continuing with that step, the string data must first undergo data conversion so that it can be converted into a numerical value.

## 4.3 DATA CONVERSION

Using the applymap function resulted in replacing "Yes" with 1 and "No" with 0. This was done since all categorical features were specified with either "Yes" or "No." Because of this, the dataset is now prepared for the application of association rule mining, which will be covered in the subsequent section.

## 4.4 ASSOCIATION RULE MINING

Association rule mining is a process that involves discovering interesting associations and links between massive amounts of data objects. This can be accomplished by using various techniques. This rule offers insight into the number of times a specific item set is found in a dataset by providing information regarding its occurrence frequency. The apriori rule mining technique was utilized in this study so that the association rule could be mined from our dataset. In order to accommodate this, the minimum support has been set to 0.1, and the minimum threshold has been established at 0.7. A total of 1150 rules were generated for the dataset as a result of the apriori rule mining technique. The top 50 rules from our dataset are displayed in table 4-2, which can be found here.

*Table 4-2 Top 50 Rules of Our Dataset*

| S.L | Rule# | Antecedents | Consequents | Support | Confidence | Lift |
|---|---|---|---|---|---|---|
| 1 | 861 | Polyphagia, delayed healing, sudden weight loss, partial paresis | Polyuria | 0.102 | 0.981 | 1.978 |
| 2 | 912 | sudden weight loss, Polydipsia, visual blurring, partial paresis | weakness | 0.131 | 0.971 | 1.656 |
| 3 | 530 | muscle stiffness, sudden weight loss, partial paresis | weakness | 0.121 | 0.969 | 1.652 |
| 4 | 1048 | Polyphagia, muscle stiffness, visual blurring, Itching | delayed healing | 0.119 | 0.969 | 2.108 |
| 5 | 1073 | Polyuria, visual blurring, sudden weight loss, partial paresis, weakness | Polydipsia | 0.119 | 0.969 | 2.162 |
| 6 | 1074 | Polyuria, visual blurring, sudden weight loss, partial paresis, Polydipsia | weakness | 0.119 | 0.969 | 1.652 |
| 7 | 972 | muscle stiffness, Polydipsia, visual blurring, partial paresis | weakness | 0.115 | 0.968 | 1.65 |
| 8 | 1102 | Polyuria, visual blurring, Polyphagia, partial paresis, weakness | Polydipsia | 0.112 | 0.967 | 2.157 |
| 9 | 851 | Polyuria, muscle stiffness, sudden weight loss, partial paresis | weakness | 0.112 | 0.967 | 1.648 |
| 10 | 701 | delayed healing, sudden weight loss, partial paresis, Polydipsia | Polyuria | 0.112 | 0.967 | 1.948 |
| 11 | 659 | Polyuria, Polyphagia, visual blurring, sudden weight loss | Polydipsia | 0.11 | 0.966 | 2.156 |
| 12 | 929 | muscle stiffness, Polydipsia, sudden weight loss, partial paresis | weakness | 0.108 | 0.966 | 1.646 |
| 13 | 840 | Polyuria, muscle stiffness, visual blurring, sudden weight loss | weakness | 0.106 | 0.965 | 1.645 |
| 14 | 688 | Polyuria, muscle stiffness, visual blurring, sudden weight loss | Polydipsia | 0.106 | 0.965 | 2.153 |
| 15 | 1050 | Polyuria, visual blurring, sudden weight loss, Polyphagia, weakness | Polydipsia | 0.104 | 0.964 | 2.152 |
| 16 | 1087 | Polyuria, visual blurring, sudden weight loss, weakness, muscle stiffness | Polydipsia | 0.102 | 0.964 | 2.151 |
| 17 | 1088 | muscle stiffness, Polyuria, visual blurring, sudden weight loss, Polydipsia | weakness | 0.102 | 0.964 | 1.643 |
| 18 | 695 | Polyuria, Itching, sudden weight loss, partial paresis | Polydipsia | 0.102 | 0.964 | 2.151 |
| 19 | 664 | Polyuria, Polyphagia, sudden weight loss, Itching | Polydipsia | 0.102 | 0.964 | 2.151 |
| 20 | 627 | Polyuria, Polydipsia, visual blurring, sudden weight loss | weakness | 0.146 | 0.962 | 1.64 |
| 21 | 477 | muscle stiffness, partial paresis, Polydipsia | weakness | 0.142 | 0.961 | 1.638 |
| 22 | 520 | visual blurring, sudden weight loss, partial paresis | weakness | 0.138 | 0.96 | 1.637 |
| 23 | 682 | Polyuria, visual blurring, partial paresis, sudden weight loss | Polydipsia | 0.123 | 0.955 | 2.132 |
| 24 | 834 | Polyuria, visual blurring, partial paresis, sudden weight loss | weakness | 0.123 | 0.955 | 1.629 |
| 25 | 774 | Polyuria, muscle stiffness, partial paresis, Polydipsia | weakness | 0.119 | 0.954 | 1.626 |
| 26 | 751 | Polyuria, muscle stiffness, visual blurring, Polydipsia | weakness | 0.117 | 0.953 | 1.625 |
| 27 | 285 | Polyuria, visual blurring, sudden weight loss | weakness | 0.154 | 0.952 | 1.624 |
| 28 | 626 | Polyuria, weakness, visual blurring, sudden weight loss | Polydipsia | 0.146 | 0.95 | 2.12 |
| 29 | 823 | Polyuria, Polyphagia, visual blurring, sudden weight loss | weakness | 0.108 | 0.949 | 1.618 |
| 30 | 957 | muscle stiffness, Polyphagia, Polydipsia, partial paresis | weakness | 0.108 | 0.949 | 1.618 |
| 31 | 924 | Itching, weakness, sudden weight loss, partial paresis | Polydipsia | 0.104 | 0.947 | 2.114 |
| 32 | 1051 | Polyuria, visual blurring, sudden weight loss, Polyphagia, Polydipsia | weakness | 0.104 | 0.947 | 1.615 |
| 33 | 1121 | Polyuria, delayed healing, partial paresis, weakness, Itching | Polydipsia | 0.102 | 0.946 | 2.112 |
| 34 | 320 | delayed healing, sudden weight loss, partial paresis | Polyuria | 0.131 | 0.944 | 1.904 |
| 35 | 911 | sudden weight loss, weakness, visual blurring, partial paresis | Polydipsia | 0.131 | 0.944 | 2.108 |
| 36 | 308 | Polyphagia, sudden weight loss, partial paresis | Polyuria | 0.16 | 0.943 | 1.901 |
| 37 | 422 | Polyphagia, sudden weight loss, partial paresis | Polydipsia | 0.16 | 0.943 | 2.105 |
| 38 | 199 | Polyuria, visual blurring, sudden weight loss | Polydipsia | 0.152 | 0.94 | 2.099 |
| 39 | 674 | Polyuria, Polyphagia, sudden weight loss, partial paresis | Polydipsia | 0.15 | 0.94 | 2.097 |
| 40 | 678 | Polyphagia, Polydipsia, sudden weight loss, partial paresis | Polyuria | 0.15 | 0.94 | 1.894 |
| 41 | 436 | Itching, sudden weight loss, partial paresis | Polydipsia | 0.117 | 0.938 | 2.094 |
| 42 | 671 | Polyphagia, delayed healing, sudden weight loss, Polydipsia | Polyuria | 0.112 | 0.935 | 1.885 |
| 43 | 832 | Polyphagia, weakness, sudden weight loss, partial paresis | Polyuria | 0.135 | 0.933 | 1.885 |
| 44 | 429 | visual blurring, sudden weight loss, partial paresis | Polydipsia | 0.135 | 0.933 | 2.083 |

| 45 | 905 | Polyphagia, weakness, sudden weight loss, partial paresis | Polydipsia | 0.135 | 0.933 | 2.083 |
| 46 | 743 | Polyuria, weakness, visual blurring, partial paresis | Polydipsia | 0.156 | 0.931 | 2.078 |
| 47 | 849 | weakness, delayed healing, sudden weight loss, partial paresis | Polyuria | 0.104 | 0.931 | 1.877 |
| 48 | 1139 | delayed healing, Polyphagia, partial paresis, Itching, Polydipsia | Polyuria | 0.104 | 0.931 | 1.877 |
| 49 | 1055 | visual blurring, sudden weight loss, Polyphagia, weakness, Polydipsia | Polydipsia | 0.104 | 0.931 | 1.877 |
| 50 | 438 | Itching, muscle stiffness, sudden weight loss | Polydipsia | 0.102 | 0.93 | 2.075 |

## 4.6 DATA PREPROCESSING

In this part, data preprocessing methods have been employed in the following manner in preparation for the subsequent simulation that would predict early onset of diabetes. Converting the target's "Class" values to their corresponding numerical values, in other words, changing "positive" to "1" and "negative" to "0." Separating the Target (Class) feature from the rest of the 15 characteristics, and storing them. Furthermore, data normalization have been applied for the continues feature "Age". The next step, which is to determine the most important feature, will include using six distinct methods, as described in the following section.

## 4.7 IMPORTANT FEATURE SELECTION
### 4.7.1 PEARSON

Pearson The correlation method evaluates the linear relationship that exists between two characteristics and generates a value that can range anywhere from -1 to 1 to show the degree to which the two characteristics are related to one another. This value shows the degree to which the linear relationship that exists between the two characteristics can be described as a correlation. The construction of a correlation matrix requires the use of this quantity. The construction of a correlation matrix is made possible by the use of correlation. By computing the relationship that exists between each feature and the goal variable, this determines the degree to which the two features are interdependent on one another. This determines the degree to which the two features are dependent on one another. When this stage of the process is finished, the next step is to discover the attribute that has the greatest substantial influence on the variable that is being sought.

### 4.7.2 CHI-2

The chi-2 test is a type of statistical analysis that can compare the actual findings with the anticipated results of a study. The purpose of this test is to determine whether or not a discrepancy between the data that have been observed and the data that have been expected can be attributed to random variation or whether or not it can be attributed to a relationship between the variables that are the focus of investigation [42], [43].

### 4.7.3 RECURSIVE FEATURE ELIMINATION

Recursive feature elimination, or RFE, is a technique that involves choosing features that are appropriate for a model and gradually eliminating the weakest features until the desired amount of features is reached. Recursive feature elimination is usually referred to by the acronym RFE [44].

### 4.7.4 LOGISTIC REGRESSION L1

Within the area of machine learning, the use of L1 regularized logistic regression is currently considered to be standard practice. This method is applicable to a wide range of classification issues, in particular those that involve a considerable number of distinct attributes[45]. It is vital to discover a solution to a problem involving convex optimization before employing L1 regularized logistic regression since this type of problem requires it.

### 4.7.5 RANDOM FOREST
A random selection of the features and the observations from the dataset are used to build each decision tree that makes up a random forest, which can have anywhere between 400 and 2000 decision trees [46]. In a random forest, there could be anywhere between 400 and 2,000 decision trees. A random forest can have anything between 400 and 12,000 decision trees at any given time.

### 4.7.6 LIGHTGBM
LightGBM, a gradient boosting framework that prioritizes teaching using the tree-based learning approach. The LightGBM creates trees vertically, in contrast to previous techniques. Most tree-growing algorithms develop their trees horizontally[47]. This suggests that, unlike other methods, the LightGBM approach constructs trees leaf-wise rather than level-wise.

The summarized output of the six model for the important feature selection have been shown in Table 4-3.

*Table 4-3 The Summarized Output of The Six Model for The Important Feature Selection*

| SL. | Feature | Pearson | Chi-2 | RFE | Logistics | Random Forest | LightGBM | Total |
|---|---|---|---|---|---|---|---|---|
| 1 | Polyuria | True | True | True | True | True | True | 6 |
| 2 | Polydipsia | True | True | True | True | True | False | 5 |
| 3 | Gender | True | True | True | True | True | False | 5 |
| 4 | weakness | True | True | True | False | False | True | 4 |
| 5 | visual blurring | True | True | True | False | False | True | 4 |
| 6 | sudden weight loss | True | True | True | False | True | False | 4 |
| 7 | partial paresis | True | True | True | False | True | False | 4 |
| 8 | Itching | True | False | True | True | False | True | 4 |
| 9 | Irritability | True | True | True | True | False | False | 4 |
| 10 | Age | True | False | True | False | True | True | 4 |
| 11 | delayed healing | True | False | True | False | False | True | 3 |
| 12 | Polyphagia | True | True | True | False | False | False | 3 |
| 13 | Genital thrush | True | False | True | True | False | False | 3 |
| 14 | Alopecia | True | True | True | False | False | False | 3 |
| 15 | muscle stiffness | True | False | True | False | False | False | 2 |
| 16 | Obesity | True | False | True | False | False | False | 2 |

From Table 4-3, the first ten elements with a value that is more than or equal to four in total have been chosen as essential features that will be placed to models to predict early diabetes. These features will be used to determine whether or not a person will develop diabetes. These are Polyuria, Polydipsia, Gender, weakness, visual blurring, sudden weight loss, partial paresis, Itching, Irritability, and Age.

The approach of consolidating feature importance across multiple analysis methods presents several advantages. Firstly, it ensures a comprehensive evaluation by considering various perspectives—ranging from traditional methods like Pearson correlation and Recursive Feature Elimination to more advanced algorithms such as Logistic regression, Random Forest, and LightGBM. This diverse range of methodologies allows for a more nuanced understanding of feature relevance, accounting for both linear and non-linear relationships within the data. Furthermore, the selection of the top ten elements based on a cumulative count of four or more aims to balance the insights from these methods, providing a collective view that can mitigate potential biases inherent in any single technique. This approach is advantageous as it acknowledges the evolving landscape of analytical methods and adapts to newer algorithms, as demonstrated by the inclusion of LightGBM, thereby embracing more sophisticated models for feature selection.

**4.8 TRAIN TEST SPLIT**
In order to carry out the application of the models and perform the task of predicting early diabetes, the ten significant characteristics that were picked have been divided into a train set and a test set. Consequently, 80% of the dataset was used for the training process, and 20% of the dataset was utilized for the testing process.

*Table 4-4 Train Test Split Dataset Description*

| Name | Description |
|---|---|
| Train set percentage | 80% |
| Test set percentage | 20% |
| Number of Train set instances | 416 |
| Number of Test set instances | 104 |
| Total Number of Images | 510 |

**4.9 K-FOLD CROSS VALIDATION**
k-Fold cross-validation is a technique that can be used to minimize the downsides of the hold-out approach. The "test just once bottleneck" can be avoided with the use of k-Fold, which offers a fresh approach to dataset segmentation[48]. Choose k folds, the number of folds. If at all possible, the dataset should be split into k equal halves. So ought to use k-1 folds as the practice set. The remaining fold will make up the test set. Use the training set to put the model through its paces. When using cross-validation, a new model must be trained independently of the model that was trained in the previous iteration. Use the test set for validation. Maintain a record of the validation's results[49]. Steps 3-6 must be repeated K times. In this analysis, K was found to have a value of 8 for each of the 10 different models.

Cross-validation accuracy= $\frac{1}{k} \sum_{i=1}^{k} \text{Accuracy}_i$

Where:
- k is the number of folds in the cross-validation.
- $\text{Accuracy}_i$ represents the accuracy of the model on the $i$th validation set or fold.

This equation sums up the accuracy obtained on each validation set and divides it by the total number of folds to calculate the average accuracy across all the validation sets.

**4.10 Machine learning technique**
**4.10.1 DECISION TREE**
An example of a decision support tool is a decision tree, which uses a tree-like model to represent decisions and the likely effects of those actions. The results of random events, resource costs, and resource utility are a few examples of these potential implications. This can be used to display an algorithm that is nothing more than a set of conditional control statements[50]. Decision trees, or more precisely decision analysis, are a common tool in the field of operations research for identifying the strategy that has the highest likelihood of success. In the area of machine learning, decision trees are a common tool.

**4.10.2 RANDOM FOREST CLASSIFIER**
A classification method made up of numerous separate decision trees is known as a random forest. It uses bagging and feature randomness when constructing each individual tree in an effort to create an uncorrelated forest of trees whose forecast by committee is more accurate than that of any individual tree. In doing so, it is hoped that a more accurate prognosis will be made.

### 4.10.3 SUPPORT VECTOR MACHINE
One of the most popular and adaptable supervised machine learning methods is the support vector machine (SVM). With its aid, activities involving classification and regression can both be completed. However, the categorization task will be the focus of debate in this thread. It is typically seen as being optimal for medium- and small-sized data sets. The main objective of the support vector machine is to find the optimum hyperplane that linearly splits the data points into two components while also maximizing the margin (SVM).

### 4.10.4 XGBOOST
The gradient boosted trees method is frequently applied and may be found in a piece of open-source software called XGBoost, where it is implemented effectively. The technique of supervised learning known as "gradient boosting" combines the predictions of a variety of less robust and more straightforward models in an effort to make an accurate forecast of a target variable. When gradient boosting is used for regression, regression trees take on the role of the weak learners. Each of these trees associates each point of input data with a leaf that stores a continuous score. XGBoost is able to minimize a regularized (L1 and L2) objective function by utilizing a convex loss function, which is based on the difference between the predicted and target outputs, and a penalty term for the complexity of the model (in other words, the regression tree functions). The training procedure is carried out in an iterative manner by adding new trees to the mix that make predictions about the residuals or errors produced by earlier trees. These new trees are then combined with the older trees to form the final prediction. The method is known as gradient boosting, and its name comes from the fact that it reduces the amount of information that is lost as additional models are added.

### 4.10.5 K- NEAREST NEIGHBOR
The supervised machine learning method known as the k-nearest neighbors (KNN) algorithm is simple and can be used for both classification and regression problems. It is simple to construct and understand, but it has a significant drawback in that it gets noticeably slower as the amount of data being used grows.

### 4.10.6 GAUSSIAN NAÏVE BAYES
The Gaussian Naive Bayes algorithm is an illustration of a probabilistic classification technique. This strategy is based on the Bayes theorem and strict independence presumptions.

$$P(a_i \mid b) = \frac{1}{\sqrt{2\pi\sigma_b^2}} \exp\left(-\frac{(a_i - \mu_b)^2}{2\sigma_b^2}\right)$$

Occasionally, suppose difference
- is independent of $b$ (i.e., $\sigma i$),
- or independent of $ai$ (i.e., $\sigma k$)
- or both (i.e., $\sigma$)

Making the assumption that the continuous values associated with each class are distributed according to a normal (or Gaussian) distribution while working with continuous data is a typical approach. In order to make working with continuous data easier, this is done. Assuming the following about the likelihood of the qualities:
According to the Gaussian Naive Bayes approach, continuous valued features and models are individually thought to correspond to a Gaussian distribution (also known as a normal distribution).

### 4.10.7 ADABOOST CLASSIFIER

An AdaBoost [51] classifier is a meta-estimator that operates by first fitting a classifier on the initial dataset, and then fitting additional copies of the classifier on the same dataset with the weights of incorrectly classified instances adjusted in a way that causes subsequent classifiers to concentrate more on difficult cases. The repetition of this procedure results in the desired accuracy. Up until the required level of categorization accuracy, this process is repeated as often as required. Until the most accurate classifier is found, this process is done many times.

**4.10.8 LOGISTIC REGRESSION**
When a dependent variable is dichotomous, the proper regression analysis to use is logistic regression (binary). The logistic regression is a predictive analysis, just like all regression analyses. Researchers utilize logistic regression to describe the data and to explain the link between one dependent binary variable and one or more independent nominal, ordinal, interval, or ratio-level variables.

**4.10.9 GRADIENT BOOSTING CLASSIFIER**
The fields of regression and classification are just two examples of potential applications for the machine learning approach known as "gradient boosting." Other potential applications include many more. It provides a prediction model in the form of an ensemble of straightforward prediction models, the vast majority of which are decision trees. The accuracy of models of this type is typically considered to be somewhat lacking in common consensus. [55][56] Gradient-boosted trees is the name of the algorithm that is generated when a decision tree is the weak learner; it routinely performs better than random forest does. Gradient-boosted trees are formed when a decision tree is the weak learner. When a decision tree is determined to be a poor learner, a gradient-boosted tree will be constructed. [55] [56] [57] It is developed in a stage-by-stage manner similar to how other boosting methods are produced; however, it generalizes the other approaches by allowing optimization of an arbitrary differentiable loss function. It is created in the same fashion as the other boosting methods. A gradient-boosted trees model is produced as a consequence of carrying out this procedure.

**4.10.10 EXTRATREES CLASSIFIER**
The program known as Train With AutoML is an application that puts into action an approach to ensemble supervised machine learning known as additional trees (short for excessively randomized trees). Excessively randomized trees is what the term "extra trees" refers to in its longer form. This method is also sometimes referred to by the term "extremely randomized trees," which is a shortcut for the longer phrase "highly randomized trees." This tactic makes use of decision trees, and the strategy's shorter term, "additional trees," alludes to the decision trees that are implemented in the tactic.

**4.10.11 DNet**
The developed model is a hybrid architecture that combines Convolutional Neural Network (CNN) and Long Short-Term Memory (LSTM) layers for effective feature extraction and sequential learning. The initial convolutional block captures key features, followed by a residual block with skip connections to enhance information flow. The model employs Batch Normalization and Dropout for regularization. The LSTM layer then captures temporal dependencies in the data.

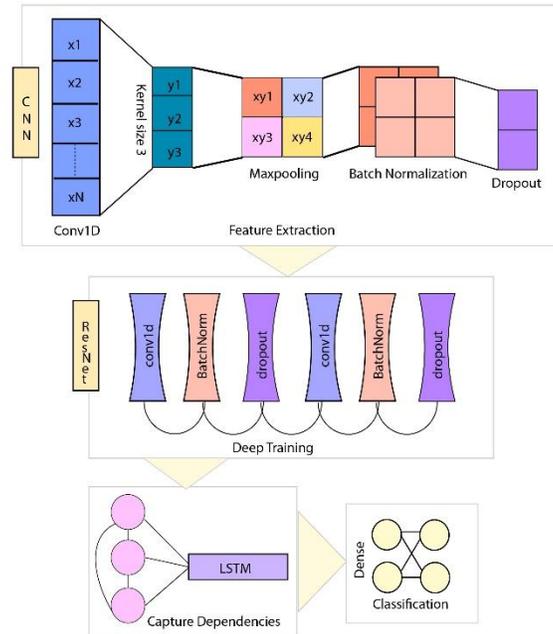

Finally, dense layers are utilized for further abstraction, culminating in a sigmoid output for binary classification. This model is designed for tasks requiring both spatial and sequential pattern recognition, displaying the versatility of combined CNN and LSTM architectures.

| **Algorithm 1: Hybrid CNN-ResNet with LSTM Model for Diabetes Prediction** |
|---|
| **Start** |
| **Define** `lr_schedule(epoch)` as the learning rate schedule function: |
|     **Initialize** `initial_learning_rate` to 0.01. |
|     **Return** `initial_learning_rate *0.9** epoch`. |
| **Define** `build_cnn_resnet_model(input_shape)` as the CNN-ResNet model building function: |
|     **Inputs**: `input_shape`, the shape of the input data. |
|     **Outputs**: `model`, the CNN-ResNet model. |
| \\**Initialize** the model architecture: |
|     **Create input layer** with shape `input_shape`: |
|         `inputs = Input(shape=input_shape)` |
| \\**Initial Convolutional Block** |
|     Apply `Conv1D` layer with 64 filters, kernel size of 3, and ReLU activation to `inputs`, producing `x`. |
|     Apply `MaxPooling1D` with pool size 2 to `x`. |
|     Apply `BatchNormalization` to `x`. |
|     Apply `Dropout` with rate 0.5 to `x`. |
| \\**Residual Block** |
|     **For each layer** in the residual block: |
|         Apply `Conv1D` with 64 filters, kernel size of 3, ReLU activation, and 'same' padding to `x`, producing `resnet_block`. |

```
        Apply BatchNormalization to resnet_block.
        Apply Dropout with rate 0.5 to resnet_block.
    If skip connection is required:
        Combine x and resnet_block using Add, producing updated x.
    Else:
        Set x to resnet_block without adding.
\\LSTM Layer
    If sequential information is to be captured:
        Apply LSTM layer with 100 units and ReLU activation to x.
\\Dense Layers
    Apply Dense layer with 50 units and ReLU activation to x.
    Apply final Dense layer with 1 unit and sigmoid activation to x, producing outputs.
\\Create and Return Model
    Define model = Model(inputs=inputs, outputs=outputs).
    Return model.
```

The algorithm builds a CNN-ResNet-LSTM model for diabetes prediction. It starts with a learning rate schedule, initialized at 0.01 and decaying by 10% each epoch. The model consists of an input layer, a convolutional block for feature extraction with dropout for regularization, followed by a residual block with skip connections for enhanced information flow. An LSTM layer captures temporal dependencies, and dense layers refine the output, with a final sigmoid layer for binary prediction. This workflow combines CNN, ResNet, and LSTM components to optimize learning efficiency and prediction accuracy.

## 4.11 PERFORMANCE METRICS

Four performance metrics that are widely used throughout this study: Accuracy, Precision, Recall, F1-Score (1-4).

$$A = \frac{TP + TN}{TP + TN + FP + FN} \quad \text{---------------------- (1)}$$

$$P = \frac{TP}{TP + FP} \quad \text{--------------------------- (2)}$$

$$R = \frac{TP}{TP + FN} \quad \text{--------------------------- (3)}$$

$$F1 - \text{Score} = 2 \times \frac{P \times R}{P + R} \quad \text{------------------ (4)}$$

where, respectively, *TP*, *TN*, *FP*, *FN*, and *FPFN* stand for true positive, true negative, false positive, and false negative. Similar to P , R , and A , P , R , and A , respectively, stand for Precision, Recall, and Accuracy[5].

## 5. RESULTS ANALYSIS

According to the results of the simulation, the Logistic Regression model achieved an accuracy of 95.19%, a Cross Val Accuracy of 91.59%, a Precision of 94.03%, a Recall of 98.44%, and an F1 Score of 96.18%. After that, the Random Forest model attained an accuracy of 98.80% percent, a cross-validation accuracy of 97.36 percent, precision of 98.44% percent, recall of 98.18 percent, and F1 Score of 99.90% percent. Following that, the model SVM attained an accuracy of 95.19%, a Cross Val Accuracy of 90.63%, a Precision of 94.03%, a Recall of 98.44%, and an F1 Score of 96.18%. Then The model KNN achieved Accuracy of 99.04%, Cross Val Accuracy of 96.15%, Precision of 98.46%, Recall of 99.00% and F1 Score of 99.22%. Then The model Decision Tree achieved Accuracy of 95.19%, Cross Val Accuracy of 93.99%,

Precision of 94.03%, Recall of 98.44% and F1 Score of 96.18%. Following that, the model AdaBoostClassifier accomplished an accuracy of 98.08%, a Cross Val Accuracy of 89.90%, a Precision of 96.97%, a Recall of 98.00%, and an F1 Score of 98.46%. Following that, the Naive Bayes GB model accomplished an accuracy of 89.42%, a Cross Val Accuracy of 88.22%, a Precision of 87.32%, a Recall of 96.88%, and an F1 Score of 91.85%. Then The model Gradient Boosting Classifier achieved Accuracy of 98.08%, Cross Val Accuracy of 96.63%, Precision of 96.97%, Recall of 97.03% and F1 Score of 98.46%. Then The model Extra Trees Classifier achieved Accuracy of 99.03%, Cross Val Accuracy of 95.19%, Precision of 97.85%, Recall of 99.87.00% and F1 Score of 99.90%. After that, the model XGB got an accuracy score of 89.42%, a Cross Val Accuracy score of 87.74%, a Precision score of 85.33%, a Recall score of 98.93%, and an F1 Score of 92.09%. This study applied hybrid model DNet and got an accuracy 99.79%, Cross Val Accuracy score 97.19%, Precision score of 98.46%, Recall 99.87%, and F1 Score 99.22%. The illustration of the models performance presented in figure 5-1.

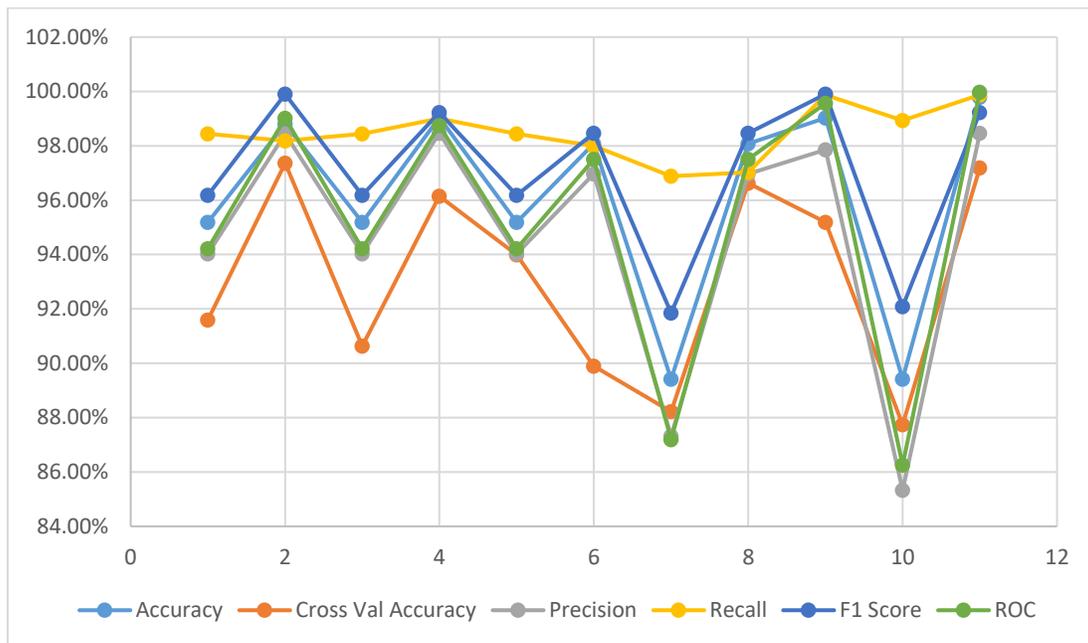

*Figure 5-1 The Illustrative Example of How Well the Model Performed. The accuracy is represented by the color blue, the accuracy of the cross-validation is represented by the color orange, the precision is represented by the color gray, the recall is represented by the color yellow, and the F1 score is represented by the color deep blue*

Figure 5-1 demonstrates that the model with the highest performance is DNet. This model has an an accuracy 99.79%, Cross Val Accuracy score 97.19%, Precision score of 98.46%, Recall 99.87%, and F1 Score 99.22%. Despite the fact that Extra Trees has also achieved nearly the same level of success as Extra Trees Classifier, with value accuracy of 99.03%, accuracy in cross-validation of 95.19%, precision of 97.85%, recall of 99.87 %, and F1 Score of 99.90%. Table 5-1 provides an overview of the overall performance of all of the models.

*Table 5-1 The Summarized Performance of All Models*

| Model | Accuracy | Cross Val Accuracy | Precision | Recall | F1 Score | ROC |
|---|---|---|---|---|---|---|
| LR | 95.19% | 91.59% | 94.03% | 98.44% | 96.18% | 94.22% |
| RF | 98.80% | 97.36% | 98.44% | 98.18% | 99.90% | 99.02% |
| SVM | 95.19% | 90.63% | 94.03% | 98.44% | 96.18% | 94.22% |
| KNN | 99.04% | 96.15% | 98.46% | 99.00% | 99.22% | 98.75% |
| DT | 95.19% | 93.99% | 94.03% | 98.44% | 96.18% | 94.22% |
| AdaBoost | 98.08% | 89.90% | 96.97% | 98.00% | 98.46% | 97.50% |
| Naïve Bayes GB | 89.42% | 88.22% | 87.32% | 96.88% | 91.85% | 87.19% |
| GB Classifier | 98.08% | 96.63% | 96.97% | 97.03% | 98.46% | 97.50% |
| Extra Trees | 99.03% | 95.19% | 97.85% | 99.87% | 99.90% | 99.57% |
| XGB | 89.42% | 87.74% | 85.33% | 98.93% | 92.09% | 86.25% |
| DNet | 99.79% | 97.19% | 98.46% | 99.87% | 99.22% | 99.98% |

Following this, a brief discussion on the confusion matrix as well as the roc for each of the applied models will be presented in the following part.

## 6. DISCUSSION

Figure 6-1 is a graphical representation of the confusion matrix that describes the anticipated results of an experiment. In addition to that, the ROC-AUC assessments of the model's performance have been shown in this study. A receiver operating characteristic curve, often known as a ROC curve, is the name given to a curve that plots the true positive rate on the ordinate of the graph and the false positive rate on the abscissa. It is the result of several different boundary values being combined into one. The area under the ROC curve, also known as AUC, is a measure of the likelihood that the computed score of the positive sample will be higher than the calculated value of the negative sample. When samples are selected at random, it is possible to investigate both the benefits and drawbacks of using the prediction model. Figure 6-1, which once more depicts the ROC curve for an experiment, reveals that the average AUC value for our model Extra Trees Classifier is 99.96%.

### 5.1 Performance evaluation

Table is a graphical representation of the confusion matrix that describes the anticipated results of an experiment. In addition to that, the ROC-AUC assessments of the model's performance have been shown in this study. A receiver operating characteristic curve, often known as a ROC curve, is the name given to a curve that plots the true positive rate on the ordinate of the graph and the false positive rate on the abscissa. It is the result of several different boundary values being combined into one. The area under the ROC curve, also known as AUC, is a measure of the likelihood that the computed score of the positive sample will be higher than the calculated value of the negative sample. When samples are selected at random, it is possible to investigate both the benefits and drawbacks of using the prediction model. Figure 6-1, which once more depicts the ROC curve for an experiment, reveals that the average AUC value for our model Extra Trees Classifier is 99.96%.

**Table. Performance comparison based on confusion matrix**

| Model | True Negative (%) | False Positive (%) | False Negative (%) | True Positive (%) |
|---|---|---|---|---|
| LR | 36 (34.62%) | 4 (3.85%) | 1 (0.96%) | 63 (60.58%) |
| RF | 40 (38.46%) | 0 (0.00%) | 0 (0.00%) | 64 (61.54%) |
| SVM | 36 (34.62%) | 4 (3.85%) | 1 (0.96%) | 63 (60.58%) |
| KNN | 39 (37.50%) | 1 (0.96%) | 0 (0.00%) | 64 (61.54%) |
| DT | 36 (34.62%) | 4 (3.85%) | 1 (0.96%) | 63 (60.58%) |
| AdaBoost | 38 (36.54%) | 2 (1.92%) | 0 (0.00%) | 64 (61.54%) |
| Naïve Bayes GB | 31 (29.81%) | 9 (8.65%) | 2 (1.92%) | 62 (59.62%) |
| Gradient Boosting Classifier | 38 (36.54%) | 2 (1.92%) | 0 (0.00%) | 64 (61.54%) |
| Extra Trees Classifier | 40 (38.46%) | 0 (0.00%) | 0 (0.00%) | 64 (61.54%) |
| XGB Classifier | 29 (27.88%) | 11 (10.58%) | 0 (0.00%) | 64 (61.54%) |
| **DNet** | **150 (93.75%)** | **10 (6.25%)** | **7 (2.73%)** | **249 (97.27%)** |

This study assessed multiple machine learning models for diabetes prediction, focusing on accuracy and reliability. Random Forest and Extra Trees Classifiers excelled, achieving high accuracy with no misclassifications, showing strong reliability. KNN and AdaBoost performed well with minimal errors, while Logistic Regression, SVM, and Decision Tree models had moderate reliability due to some misclassifications. Ensemble methods like Gradient Boosting showed enhanced performance, though Naïve Bayes GB and XGBoost had higher false positive rates, limiting their medical applicability. The D-Net model, a hybrid CNN-LSTM architecture, achieved the best results with high true positive and true negative rates, highlighting its potential as an accurate and reliable model for diabetes prediction in healthcare applications.

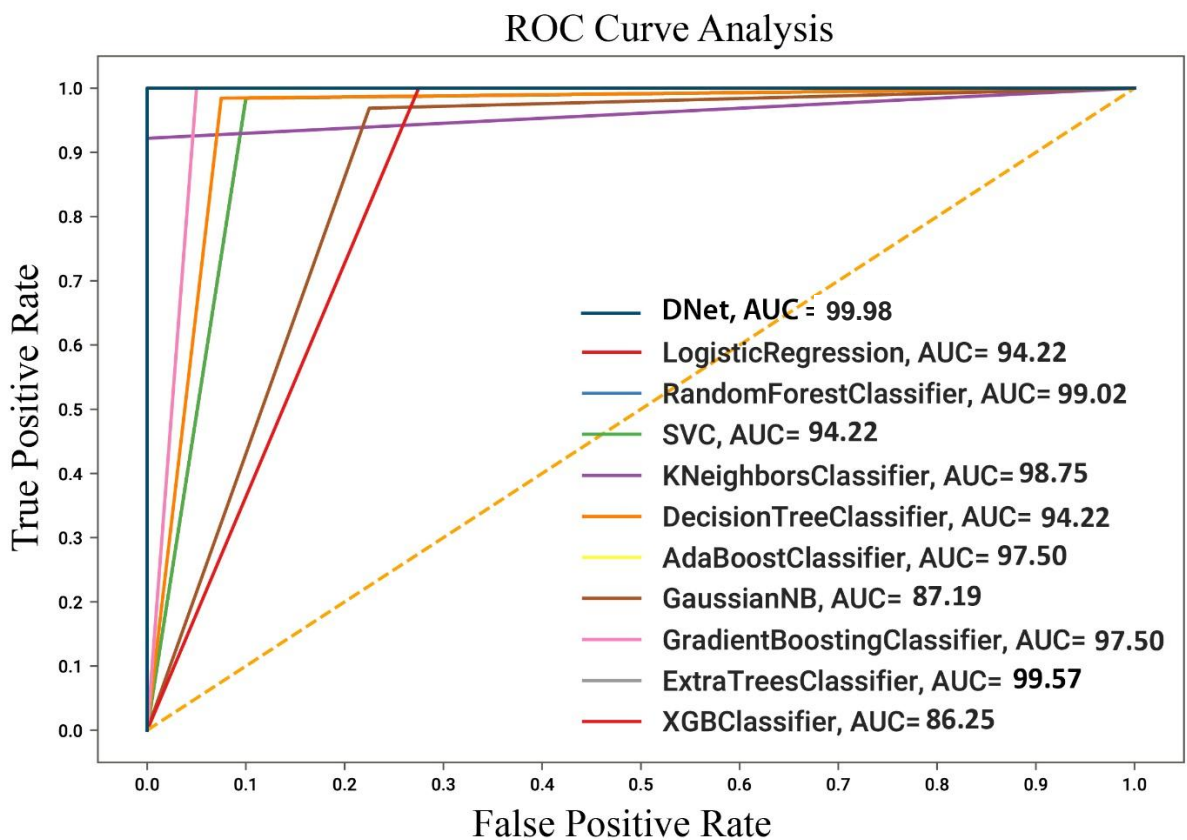

*Figure 6-2 The ROC Curve for the Experiment*

Figure 6-2 demonstrates that the Logistic Regression model was successful in achieving a ROC value of 94.22%. The next step was for the Random Forest model to reach a ROC value of 99.02 percent. After that, the ROC value for the SVM model was found to be 94.22%, while the ROC value for the KNN model was found to be 98.75%. The next model, Decision Tree, managed to get a ROC value of 94.22%. Following that, the AdaBoost Classifier model achieved a ROC value of 97.50%, and the Naive Bayes GB model achieved a ROC value of 87.19%. The subsequent Gradient Boosting Classifier model acquired a ROC value of 97.50%. The subsequent model, Extra Trees Classifier, attained a ROC value of 99.57%. The subsequent XGBClassifier model reached a ROC score of 86.25 % and the subsequent model DNet model reached a ROC score of 99.98%.

The work that has already been done and the model this study has suggested will be compared in the part that follows.

## 7. COMPARE WITH THE EXISTING WORK

The comparative analysis of our proposed Early Stage Diabetics Risk Prediction model against existing studies presents a comprehensive understanding of the performance of various models on distinct datasets. The existing works, encompassing a variety of machine learning algorithms such as Boosted Regression, Logistic Regression, Naïve Bayes, Support Vector Machines, Random Forest, Hidden Markov Models, Decision Trees, Neural Networks, among others, were applied to datasets including the Pima Indian Diabetes Dataset, hospital physical examination data, and global datasets. The reported accuracies across these studies ranged between 65.38% and 90.91%. This wide range suggests the diversity in methodologies and datasets used, impacting the predictive performance. Tripathi et al.[20] used a model known as Boosted Regression, and they were able to achieve an accuracy of 90.91% by doing so. Next, Alaa Khaleel et al. [21] used the Logistic Regression (LR), Naive Bayes (NB), and K-nearest Neighbor (KNN) models, and they were successful in achieving accuracy of 94% for the LR model, 79% for the NB model, and 69% for the KNN model. After that Zou et al. [22]The accuracy level that was attained using the Random Forest Can technique was 80.84%. Sisodia et al. are next. [24] used the Naive Bayes technique and was able to get an accuracy of 76.30 percent. The accuracy was 83.20 percent when Ramesh et al. [25] utilized the Support Vector Machine after that. Perveen et al. are next. [26] was able to achieve an accuracy of 86.9% using the Hidden Markov Model. Following are Kavakitis et al. [28] The Support Vector Machine was used, and an accuracy success rate of 85% was achieved. Yahyaoui et al. [30] are next. The accuracy for SVM, Random Forest (RF), and DL was 83.67% for SVM, 76.81% for RF, and 65.38% for DL in, which used these methods. Next, Sonar et al. [31] used the Decision Tree, the Naive Bayes model, and the Support Vector Machine and was able to get accuracy levels of 85% for the Decision Tree, 77% for the Naive Bayes model, and 77.3% for the Support Vector Machine. Next, using Random Forest (RF) and Support Vector Machines (SVM), Sivaranjani et al. [32] achieved accuracy of 83.3% for RF and 81.4% for SVM, respectively. After that, Saha et al. [33] used a neural network and were able to achieve an accuracy of 80.4%. The Random Forest algorithm and the Naive Base Method were then used by Pavani et al. [35], and they both produced results with an accuracy of 80%. This is a description of both the Random Forest approach and the Naive Base Method. While several studies showcased promising accuracies, none surpassed the 90.91%. Sadhu and Jadli [52] explored Logistic Regression (LR), Naive Bayes (NB), Random Forest (RF), Multi-layer Perceptron (MLP), Decision Trees (DT), k-Nearest Neighbors (KNN), and Support Vector Machines (SVM), identifying Random Forest as the best-performing model with an accuracy of 98%. Alpan and Ilgi [53] delved into LR, NB, RF, Regression Trees (RT), DT, KNN, and SVM, ultimately favoring k-Nearest Neighbors (KNN) with an accuracy of 98%. Xue et al. [54] focused on Support Vector Machines (SVM), Naive Bayes (NB), and LightGBM, with SVM emerging as the optimal model boasting an accuracy of 98.7%. mark until our proposed model utilizing the UCI Machine Learning Repository dataset and employing the Extra Tree Classifier algorithm. This model achieved a substantially higher accuracy of 99.96%, indicating a remarkable improvement in predictive capability compared to existing methodologies.

*Table 7-1 The summarized comparison of the Early Stage Diabetics Risk Prediction model's performance (UCI machine learning repository)*

| Method | Machine learning algorithms | Best model | Accurate |
|---|---|---|---|
| Sadhu and Jadli [53] | LR, NB, RF, MLP, DT, KNN, SVM | RF | 0.98 |
| Alpan and Ilgi [54] | LR, NB, RF, RT, DT, KNN, SVM | KNN | 0.98 |
| Xue et al. [55] | SVM, NB, LightGBM | SVM | 0.987 |
| **Our Proposed** | **LR, RF, SVM, KNN, DT, NB, GBC, Extra tree, XGB, AdaBoost, DNet** | **DNet** | **99.79** |

*Table 7-2 The summarized comparison of the Early Stage Diabetics Risk Prediction model's performance (Others Dataset)*

| Method | Dataset Name | Description | Accurate |
|---|---|---|---|
| Tripathi et al,. [20] | Pima Indian Diabetes Dataset from the Kaggle ML repository | Boosted Regression model | 90.91% |
| Alaa Khaleel et al,. [21] | PIDD-Pima Indians Diabetes Dataset | Logistic Regression (LR) | 94% |
| | | Naïve Bayes (NB) | 79% |
| | | K-nearest Neighbor (KNN) | 69% |
| Zou et al,. [22] | PIDD-Pima Indians Diabetes Dataset | Random Forest Could | 80.84% |
| Sisodia et al,. [24] | The hospital physical examination data in Luzhou, China | Naive Bayes' | 76.30% |
| Ramesh et al,. [25] | PIDD-Pima Indians Diabetes Dataset | Support Vector Machine | 83.20% |
| Perveen et al,. [26] | PIMA Indian Diabetes Database | Hidden Markov Model | 86.9% |
| Kavakiotis et al,. [28] | accessed from the University of California, Irvine ML repository | SVM | 85% |
| Yahyaoui et al,. [30] | Canadian Primary Care Sentinel Surveillance Network (CPCSSN) | SVM | 83.67% |
| | | RF | 76.81% |
| | | DL | 65.38% |
| Sonar et al,. [31] | PIDD-Pima Indians Diabetes Dataset | Decision Tree | 85% |
| | | Naïve Bayes | 77% |
| | | Support Vector Machine | 77.3% |
| Sivaranjani et al,. [32] | Pima Indian Diabetes Dataset from the Kaggle ML repository | Random Forest (RF) | 83% |
| | | SVM | 81.4%. |
| Saha et al,. [33] | PIDD-Pima Indians Diabetes Dataset | Neural Network | 80.4% |
| Pavani et al,. [35] | Global data set | Random Forest algorithm | 80% |
| | | Naive Base Method | 80% |
| **Our Proposed** | UCI Machine Learning Repository Dataset | **DNet** | **99.79%** |

## 8. SUMMARY

In order to create an accurate simulation of the dataset that was used for this study, data preparation techniques such as data conversion and data normalization were utilized. In addition, association rule mining was utilized in order to identify the typical manifestations of diabetic symptoms. After that, a total of six distinct methods were utilized in order to arrive at the final decision about the dataset's primary attribute. Following that, eleven distinct models were applied to the previously selected and highlighted dataset. This study compared the results that each model generated for us so that this study could identify which one had given us the best overall results. According to the performance matrices, the DNet Model achieved the highest possible level of performance, achieving a perfect score in every category 99.79% accuracy. It is possible for us to declare that our DNet model even performs better than the work that is currently accessible. This study offers clinical physicians something new as well as something that can be of assistance to them. The lack of availability of larger databases was the primary challenge that this study faced with our efforts. Nonetheless, in order to optimize a model to its best potential, one needs initially have access to a substantial dataset. In future research, this study plans to delve deeper into diabetic symptom manifestations and optimize the predictive model further. The goal is to refine the DNet, which demonstrated exceptional performance, by leveraging advanced machine learning techniques. Acquiring access to larger and more diverse datasets remains a primary objective, aiming to uncover subtle patterns within diabetic symptoms for more precise predictions. The study also emphasizes the importance of interpretability, aiming to make the model's decision-making process more understandable for clinicians. Embracing emerging technologies and integrating patient-reported data can provide a holistic approach for enhancing diabetes diagnosis and management.

**ABBREVIATION**
AI = Artificial Intelligence
ANN = Artificial neural network
AUC = Area under the ROC Curve.
CART = Classification and Regression Tree
CNN = Convolutional Neural Networks
DM = Diabetes
DSS = Decision Support System
DT = Decision Tree
EHRs = Electronic Health Records
EDA= Exploratory Data Analysis
FDRSM = Framingham Diabetes Risk Scoring Model
GBC= Gradient Boosting Classifier
HMM= Hidden Markov Model
IDA = International Diabetes Federation
KNN = K-Nearest Neighbors
LDA = Linear Discriminant Analysis
LR = Logistic Regression
ML = Machine Learning
MRMR = Maximum Relevance — Minimum Redundancy
MLP = Multilayer Perceptron
NB = Naive Bayes
PIMA = Pima Indian Diabetes Database
PCA = Principal Component Analysis
PCA = Principal Component Analysis
RF = Random Forest
ROC = Receiver Operating Characteristic Curve
SVM = Support Vector Machine
UCI = University of California Irvine
WHO = World Health Organization
WEKA = Waikato Environment for Knowledge Analysis